\documentclass[journal]{IEEEtran}
\ifCLASSINFOpdf
\else
\fi
\usepackage[numbers,sort&compress]{natbib}
\usepackage{hyperref}       
\usepackage{url}            
\usepackage{booktabs}       
\usepackage{amsfonts}       
\usepackage{nicefrac}       
\usepackage{microtype}      
\usepackage{amsmath}
\usepackage{cleveref}
\usepackage{graphicx}
\usepackage{amssymb}
\usepackage{subcaption}
\usepackage{caption}
\begin{document}
%
\title{Retinal Vessel Segmentation under Extreme Low Annotation: A Generative Adversarial Network  Approach}
%
%
%

\author{Avisek~Lahiri*,
        Vineet~Jain*, Arnab~Mondal*,
        and~Prabir~Kumar~Biswas,~\IEEEmembership{Senior~Member,~IEEE}
\thanks{* Equal contribution (Order decided by dice roll)}
\thanks{Authors are with Dept. of E\&ECE, Indian Institute of Technology Kharagpur, India. All correspondence to Avisek Lahiri,  e-mail: avisek@ece.iitkgp.ernet.in}
\thanks{This work has been submitted to the IEEE for possible publication. Copyright may be transferred without notice, after which this version may no longer be accessible.}
}

%
%

\markboth{}%
{Shell \MakeLowercase{\textit{et al.}}: Bare Demo of IEEEtran.cls for IEEE Journals}
%



\maketitle

\begin{abstract}
Contemporary deep learning based medical image segmentation algorithms require hours of annotation labor by domain experts. These data hungry deep models perform sub-optimally in the presence of limited amount of labeled data. In this paper, we present a data efficient learning framework using the recent concept of Generative Adversarial Networks; this allows a deep neural network to perform significantly better than its fully supervised counterpart in low annotation regime. The proposed method is an extension of our previous work with the addition of a new unsupervised adversarial loss and a structured prediction based architecture. To the best of our knowledge, this work is the first demonstration of an adversarial framework based structured prediction model for medical image segmentation. Though generic, we apply our method for segmentation of blood vessels in retinal fundus images. We experiment with extreme low annotation budget (0.8 - 1.6\% of contemporary annotation size). On DRIVE and STARE datasets, the proposed method outperforms our previous method and other fully supervised benchmark models by significant margins especially with very low number of annotated examples.  In addition, our systematic ablation studies suggest some key recipes for successfully training GAN based semi-supervised algorithms with an encoder-decoder style network architecture.
\end{abstract}


%
\IEEEpeerreviewmaketitle

\section{Introduction}
With the relatively new breakthrough in large scale object recognition by Krizhevsky \textit{et al.,}\cite{krizhevsky2012imagenet}, Convolutional Neural Networks (CNN) and `deep learning'(DL) have achieved unprecedented success in numerous computer vision applications such as object detection\cite{ren2015faster}, semantic segmentation \cite{noh2015learning}, video understanding \cite{yao2015describing}, visual question-answering \cite{antol2015vqa} to list a few. Inspired by the flexibility of CNNs to adapt to novel computer vision problems, a recent surge of interest has been instigated among the medical image processing community to leverage the rich feature learning and representation prowess of CNNs. In recent years, CNNs have been applied in numerous medical image and video understanding pipelines such as segmenting areas of interest from medical images \cite{pereira2016brain,moeskops2016automatic, naylor2017nuclei,unet, first} and sequences \cite{lin2016inference}, medical video understanding \cite{twinanda2017endonet}, reconstruction \cite{jin2017deep}, anomaly region detection \cite{dou2016automatic,van2016fast,roth2014new}. The list is by no means exhaustive; readers are encouraged to refer to \cite{survey} for a detailed survey on applications of DL in medical image analysis.
\par However the success of DL comes at a price. CNN models are significantly complex with millions of trainable parameters. For example, popular architectures such as AlexNet \cite{krizhevsky2012imagenet} and VGG-Net\cite{vgg} have 60 million and 138 million parameters respectively. Such gigantic deep architectures easily overfit on small training datasets with low training error but manifests high test error. Curating manually annotated dataset is both time consuming and costly. Even though for natural computer vision problems the current trend is to annotate large scale data with mechanical turks \cite{buhrmester2011amazon}, annotating medical data often requires domain specific experts. This instigates the need for methods to train CNNs with limited amount of annotated data. Recent regularization techniques such as dropout \cite{dropout} and batch normalization \cite{ioffe2015batch} have shown promise in preventing over fitting; however a small dataset with regularization during training can easily lead to under fitting, wherein, during the training phase itself, a CNN is unable to approximate the input to output functional mapping appreciably, thereby manifesting high error rates on both training and testing data. Fine-tuning a pre-trained CNN (in most cases pre-trained for object recognition on ImageNet) for specific medical imaging tasks \cite{tajbakhsh2016convolutional} is the current trend to train a CNN with limited annotated data. Though promising, fine-tuning methods train a CNN by only annotating a fraction of available data, while the remaining unannotated data remains unused.
\par In this paper, we try to address the central question: \textit{`Can we learn from both unannotated and annotated data?'} We build upon our previous work\cite{cvpr} on semi-supervised learning which leverages the use of Generative Adversarial Networks (GAN) \cite{goodfellow2014generative}. To demonstrate the effectiveness of the proposed method, we select a specific application - the task of segmenting blood vessels in retinal fundus images. The motivation of this paper is not to present yet another supervised deep model for retinal vessel segmentation. The main objective is to perform segmentation using as little as 0.8-1.6\% of annotation samples used by contemporary deep models. For example, methods of \cite{avijit,lahiri2016deep,tmi} all use around 60,000 training patches, whereas in this work we aim to utlizie only 500-1000 annotations. Proposed framework is completely generic and our findings and recommendations can be embraced for any other medical image domain in which the the aim is to perform some discriminative task with limited amount of labeled data but plethora of unlabeled data. Our contributions in the paper can be summarized as:\\
\begin{figure*}[!t]
\centering
\includegraphics[scale = 0.4]{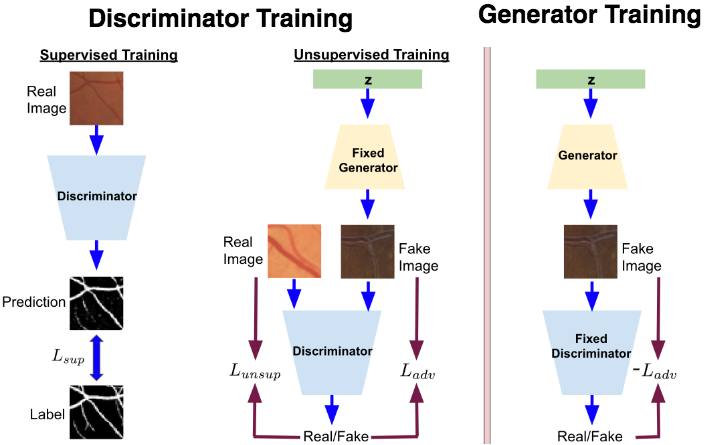}
\caption{Visualization of training of discriminator and generator networks for proposed semi-supervised learning. Here, we show the vanilla version of generator training as presented in \cite{goodfellow2014generative}, wherein, the generator is trained to create images to maximize its log likelihood (by minimizing $-L_{adv}: Eq. \ref{eq_adversarial}$) of belongingness to real class as determined by the discriminator. Discriminator training consists of two parts: (a) With limited amount of labeled examples, given an input patch, it generates a segmentation output to minimize $L_{sup}$ (Eq. \ref{eq_supervised}); (b) On unlabeled examples, discriminator predicts the domain of origin (real/fake) when fed with patches from real dataset(minimizing $L_{unsup}$: Eq. \ref{eq_unsupervised}) and fake patches (minimizing $L_{adv}$: Eq. \ref{eq_adversarial}) created by the generator. }
\label{fig_flow}
\end{figure*}
\begin{itemize}
\item We add an unsupervised loss function to our previously proposed GAN based semi-supervised framework \cite{cvpr}. This enables learning from both unlabeled and labeled data under a multitask objective setting.
\item We extend the `center-pixel'(CP) prediction framework in \cite{cvpr} to a `structured-prediction'(SP) setting by posing segmentation as \textit{multi-label} inferencing problem.
\item Several architectural and optimization recommendations (Sec. \ref{sec_gan_hacks}) are provided for successfully training a SP based semi-supervised GAN framework. To the best of our knowledge, this is the first demonstration of adversarial semi-supervised learning under SP framework for segmentation application in medical images.
\item For working with low number of annotated examples, our studies reveal  important trade-offs between number of annotations v/s diversity of annotations (Sec. \ref{sec_diversity}).
\item Evaluations (area under ROC curve) on DRIVE and STARE datasets reveal that \textit{a)} incorporation of unsupervised loss function boosts the performance of our previous method  and \textit{b)} proposed SP based method significantly outperforms current benchmark supervised models specifically with extremely low annotated examples (Sec. \ref{sec_state_of_art}).
\end{itemize}
\section{Related Work}
Traditional methods for blood vessel segmentation in fundus images can be classified into \textit{unsupervised} and \textit{supervised} paradigms. The former type of methods hard-code the local properties of structures to be detected into the algorithms. Exemplary works from this group leverage the concepts of line detectors \cite{tmi12}, co-occurrence matrix \cite{tmi13}, co-linearly aligned  difference-of-Gaussian filters \cite{tmi14} and active contour model \cite{tmi15} to list a few. In supervised methods, an image patch is corresponded to a ground truth annotation patch and a learning algorithm is deployed to learn the mapping from image space to label space. Methods of ridge features with nearest neighbor classifier \cite{tmi9}, Gabor wavelets with Bayesian classifiers \cite{tmi16}, morphological operations with Gaussian mixture model classifiers \cite{tmi17} are some of the exemplary works in this genre. A more detailed review of these traditional methods have been documented in \cite{tmi8}.
\par Liskowski and Krawiec \cite{tmi} first demonstrated the efficacy of using CNNs for retinal vessel segmentation by posing the segmentation as a two class classification problem, wherein the positive class is the blood vessel and the background or the non-vessel pixels constitute the negative class. Following \cite{tmi}, there has been a series of efforts towards DL based retina segmentation. One genre of effort \cite{lahiri2016deep,roy2015dasa} is to first pre-train deep denoising stacked autoencoder \cite{vincent2010stacked} in an unsupervised setting and then later fine-tune the model with labeled examples. Other approaches \cite{avijit,fu2016deepvessel,maninis2016deep} are analogous to the overall framework of \cite{tmi} wherein a CNN or an ensemble of CNNs are trained for vessel detection. These algorithms are usually trained on a humongous number of annotated patches. For example, \cite{tmi} was trained on 3.8$\times$10$^6$ while \cite{avijit} was trained on 1.2$\times$10$^5$ patches. Contrary to this trend, we are interested to work with as low as only 500 annotated samples.
\section{Method}
\subsection{Generative Adversarial Networks}
We begin by reviewing the concept of Generative Adversarial Networks (GAN) \cite{goodfellow2014generative}. A GAN consists of two parametrized deep neural networks, viz., generator, $G_{\theta_G}$, and discriminator, $D_{\theta_D}$. The task of the generator is to yield an image, $x\in \mathcal{R}^{H\times W \times 3}$ with a latent vector, $z\in \mathcal{R}^d$, as input. $z$ is sampled from a known distribution, $p_z(z)$. A common choice for this \cite{goodfellow2014generative} is $z\sim \mathcal{U}[-1,1]^d$. The discriminator is pitted against the generator to distinguish real samples (sampled from $p_{data}$) from fake/generated samples. Specifically, discriminator and generator play the following min-max game on $V(D_{\theta_D},G_{\theta_G})$:
\begin{equation}
\begin{aligned}
& \underset{G_{\theta_G}}{min}~~ \underset{D_{\theta_D}}{max}~~ V(D_{\theta_D}, G_{\theta_G}) = \mathbb{E}_{x\sim p_{data}(x)}[\log D_{\theta_D}(x)]\\
& ~~~~~~~+\mathbb{E}_{z\sim p_{z}(z)}[1 - D_{\theta_d}(G_{\theta_G}(z))].
\end{aligned}
\label{eq_gan_main_goodfellow}
\end{equation}
With enough capacity, on convergence, $G_{\theta_G}$ fools $D_{\theta_D}$ at random \cite{goodfellow2014generative}. In medical imaging literature, recently GANs have been used for generating patho-realistic images \cite{costa2018end, nie2018medical}, fully supervised segmentation \cite{izadi2018generative, li2017brain}, denoising \cite{wolterink2017generative, yi2018sharpness}. Our work has a different motivation; following the work in \cite{improved}, we wish to extend GANs for semi supervised learning to formulate a data efficient learning paradigm for biomedical images.
\subsection{Semi-supervised Learning using GANs}
\label{sec_semi}
In usual supervised setting, a classifier usually has $n_c$ number of output nodes while classifying an input $x$ to $n_c$ classes. Usually, it outputs a vector of unnormalized logits, {$l_1$, $l_2$, ..., $l_{n_c}$} which can be turned into normalized class probabilities with softmax operation: $p_{model}(y=k|x) = \frac{\exp(l_k)}{\sum_{i=1}^{n_c} \exp(l_i)}$ gives the probability of class label($y$) to belong to class $k$ for the input $x$. Cross entropy loss between the original class labels and predicted class distribution, $p_{model}(y|x)$ is used to train the discriminative model.
\par To perform semi-supervised learning using GANs, we need to augment one more output node to the classifier (which is also the discriminator, $D_{\theta_D}$ in our case). This extra node for class, $y$ = $n_{c} + 1$, corresponds to the class of fake/generated samples coming from the generator network, $G_{\theta_G}$. $p_{mode}(y = n_c+1|x)$ can be seen as the probability of belonging to fake class; this corresponds to the term $[1 - D_{\theta_D}(G_{\theta_G}(z))]$ in Equation \ref{eq_gan_main_goodfellow}. This enables us to update the parameters of the discriminator even with unlabeled real data by maximizing $\log p_{model}(y \in {1, 2, ..., n_c})|x$. Merging all of these, finally we have three different components in final discriminator loss $L_D(\cdot)$ to update parameters, $\theta_D$ of the discriminator;
\begin{equation}
L_D = L_{sup} + L_{adv} + L_{unsup}
\label{eq_total_loss}
\end{equation}
where,
\begin{equation}
L_{sup} = -~\mathbb{E}_{x,y\sim p_{data}(x,y)}\log p_{model}(y|x, y<n_c+1)
\label{eq_supervised}
\end{equation}
\begin{equation}
L_{adv} = -\mathbb{E}_{z\sim p_z(z)} \log p_{model}(y=n_c+1|G_{\theta_G}(z))
\label{eq_adversarial}
\end{equation}
\begin{equation}
L_{unsup} = -~ \mathbb{E}_{x\sim p_{data}(x)} \log [1 - p_{model}(y = n_c + 1|x)]
\label{eq_unsupervised}
\end{equation}
The three components of the loss function are:
\begin{itemize}
\item $L_{sup}$ is the usual cross entropy loss in which the target is to maximize the predicted probability over the correct class label.
\item $L_{adv}$ encourages the classifier to place high probability for fake class when the input is a fake/generated sample.
\item $L_{unsup}$ penalizes the classifier if it gives high probability to fake class when input is a real unlabeled sample.
\end{itemize}
It is to be noted that the formulation in our previous work \cite{cvpr} did not include the $L_{unsup}$ component and thus it was not possible to learn the parameters of the discriminator/classifier on the large amount of unlabeled real data. The two tasks of assigning class label to an image and determining whether the image is real/fake shares some low level feature representation commonalities such as identification of textures, structural regularities. Thus setting up Equations \ref{eq_supervised} and \ref{eq_unsupervised} in a multi task learning setting enables more meaningful gradient updates while training the discriminator. In Fig. \ref{fig_flow} we visualize the components of loss functions used for training the discriminator.
\subsection{Center Pixel v/s Structured Prediction}
There are two major paradigms for patchwise segmentation of medical images, namely center-pixel prediction (CP) and structured prediction (SP).
\par Let $P$ be the domain of sampled patches from the dataset such that any $p \sim P \in \mathbb{R}^{W \times W \times 1}$, where $W \times W$ is the resolution of the patches and usually $W \in 2\mathbb{N} + 1$. Also, let $Y$ be the corresponding label space for $P$, such that for a given patch, $x_p$, we have its corresponding label, $y_p \in \mathbb{R}^{W \times W }$. $y_p^{i, j}$ is the label information at location (i, j) for patch $x_p$. In case of center pixel prediction, the objective is to learn a parametrized ($
\theta _C$) functional mapping, $f_{\theta _C}: P \Rightarrow \mathbb{R}^{1 \times 1}$. Essentially this means that given an image patch, the function returns a single scalar value to predict the probability of the center pixel of that patch to belong to foreground or `vessel' class. $
\theta _C$ is optimized according to,
$$\theta_C^* = \underset{\theta_C}{\arg\min} ~-\sum_{p=1}^m y_p^{(W/2, W/2)} \log(f_{\theta _C}(x_p))$$
\begin{equation}
+ ~ (1 - y_p^{(W/2, W/2)}) \log(1 - f_{\theta _C}(x_p))
\label{eq_center_pixel}
\end{equation}
This was the procedure we followed in \cite{cvpr}.
\par In contrast, structured prediction learns a parametrized function,  $f_{\theta _S}: P \Rightarrow \mathbb{R}^{W \times W}$. $\theta_S$ is optimized as,
$$\theta_S^* = \underset{\theta_C}{\arg\min} ~-\sum_{p=1}^m \sum_{i=1}^W \sum_{j=1}^W y_p^{(i, j)} \log(f_{\theta _C}(x_p))$$
\begin{equation}
+ ~ (1 - y_p^{(i, j)}) \log(1 - f_{\theta _C}(x_p))
\label{eq_center_pixel}
\end{equation}
In this methodology, for a given image patch, the function simulteneously predicts the probability of all the pixels in the patch belonging to the `vessel' class, instead of just the center pixel.
\begin{table*}[!t]
\centering
\tiny
\caption{Network architecture of our U-Net discriminator}
\begin{tabular}{lllllllllllllllll}\hline\hline
                                                      Layer Names      & C$_1$                                                     & C$_2$    & P$_1$    & C$_3$                                                      & C$_4$    & P$_2$    & C$_5$                                                      & U$_1$                                                      & Con$_1$                                                     & C$_6$                                                      & C$_7$    & U$_2$                                                      & Con$_2$                                                     & C$_8$                                                      & C$_9$    & C$_{10}$                                                     \\\hline
Operations                                                 & \begin{tabular}[c]{@{}l@{}}Conv +\\ Droput\end{tabular} & Conv  & Pool  & \begin{tabular}[c]{@{}l@{}}Conv +\\ Droput\end{tabular} & Conv  & Pool  & \begin{tabular}[c]{@{}l@{}}Conv +\\ Droput\end{tabular} & \begin{tabular}[c]{@{}l@{}}Upsample\\ (2X)\end{tabular} & \begin{tabular}[c]{@{}l@{}}Concat\\ (U1+C4)\end{tabular} & \begin{tabular}[c]{@{}l@{}}Conv+\\ Dropout\end{tabular} & Conv  & \begin{tabular}[c]{@{}l@{}}Upsample\\ (2X)\end{tabular} & \begin{tabular}[c]{@{}l@{}}Concat\\ (U2+C2)\end{tabular} & \begin{tabular}[c]{@{}l@{}}Conv \\ +Droput\end{tabular} & Conv  & \begin{tabular}[c]{@{}l@{}}Conv+\\ Softmax\end{tabular} \\\hline
\begin{tabular}[c]{@{}l@{}}Input\\ Resolution\end{tabular} & 48X48                                                 & 48X48 & 48X48 & 24X24                                                   & 24X24 & 24X24 & 12X12                                                   & 12X12                                                   & 24X24                                                    & 24X24                                                   & 24X24 & 24X24                                                   & 48X48                                                    & 48X48                                                   & 48X48 & 48X48                                                   \\\hline
\begin{tabular}[c]{@{}l@{}}Input\\ Channels\end{tabular}   & 1                                                       & 32    & 32    & 32                                                      & 64    & 64    & 64                                                      & 64                                                      & 128                                                      & 128                                                     & 64    & 64                                                      & 96                                                       & 96                                                      & 32    & 32                                                      \\\hline
\begin{tabular}[c]{@{}l@{}}Output\\ Channels\end{tabular}  & 32                                                      & 32    & 32    & 64                                                      & 64    & 64    & 64                                                      & 64                                                      & 128                                                      & 64                                                      & 64    & 64                                                      & 96                                                       & 32                                                      & 32    & 2                                                       \\\hline
Kernel Size                                                & 3X3                                                     & 3X3   & 2X2   & 3X3                                                     & 3X3   & 2X2   & 3X3                                                     & -                                                       & -                                                        & 3X3                                                     & 3X3   & -                                                       & -                                                        & 3X3                                                     & 3X3   & 1X1    \\\hline
\end{tabular}
\label{tab_disc_architecture}
\end{table*}
\subsection{Generator Training}
In its vanilla form, the generator network can be trained by following the zero sum min-max formulation in \cite{goodfellow2014generative}, i.e., by minimizing generator loss function, $L_G$ according to,
\begin{equation}
L_G = -L_{adv}.
\end{equation}
In this objective, the generator tries to generate samples to fool the discriminator to place high log likelihood on fake samples towards real class. This vanilla formulation in shown in Fig. \ref{fig_flow}. However, for semi supervised learning under GAN setting, findings in \cite{improved} suggest `feature matching' as a preferred method over normal GAN loss for training generator. The key idea behind feature matching is that for a successful generator, the expected intermediate activations within the discriminator over a mini-batch should be same for real and fake class. This is because ultimately, it is the cascade of these intermediate representations that compel the discriminator to ascertain a given sample into real or fake class. Let, $\phi_l(\cdot) \in \mathbb{R}^{h_l, w_l, c_l}$ be an intermediate representation from layer, $l$, of discriminator; $h_l, w_l, c_l$ being the height, width and channel count of the representation. The feature mappping loss for the generator is defined as,
\begin{equation}
L_G = \psi(\mathbb{E}_{x \sim p_{data}(x)}\phi(x),~ \mathbb{E}_{z \sim p_{z}(z)}\phi(G(z))),
\label{eq_feature_matching}
\end{equation}
gwhere $\psi(\cdot)$ is any distance metric. This loss captures the expected distance between the intermediate representations $\phi_l(\cdot)$ of layer $l$ for a batch of real and fake samples. In our case we experimented with L$_2$ distance as a representation of $\psi(\cdot)$.
\subsection{Joint Training of Generator and Discriminator}
We follow the joint iterative optimization setup as in \cite{goodfellow2014generative} to simultaneously train the generator and discriminator network. For the update of discriminator, we keep the generator fixed and parameters of discriminator are updated based on $L_D$. Conversely, during generator update, we fix the discriminator and update parameters of generator based on $L_G$; $L_G$ can be either $-L_{adv}$ for vanilla version of GAN or feature matching loss as in Eq. \ref{eq_feature_matching}.
\section{Implementation Details}
\subsection{Practical Realization of Discriminator}
As discussed in Sec. \ref{sec_semi}, for a semi supervised GAN with $K$ classes, the modified discriminator needs to have $K+1$ output nodes to account for the extra `FAKE' class. However, as pointed out in \cite{improved}, having $K+1$ output nodes is an over parametrization because, subtracting a general function, $g(x)$, from each of the logits, i.e., $l_k := l_k - g(x) \forall k$, does not change the softmax evaluation. Thus we can even set the $l_{K+1}(x) = 0~ \forall x$, in which case, $L_{sup}$ becomes the usual supervised loss with $K$ nodes and the discriminator output can be written as, $D(x) = \frac{\sum_{k=k}^K \exp[l_k(x)]}{ \sum_{k=k}^K \exp[l_k(x)] + 1}$. For more details, readers are directed to \cite{improved}. This trick enables the discriminator network to be an usual deep neural network with $K$ output nodes (in case of classification) or, as in our case, $K$ output channels(each channel representing pixel wise class probability).
\par We adopt the state-of-the-art encoder-decoder based U-Net architecture as proposed in \cite{ronneberger2015u}. The U-Net model consists of an encoder section which creates a bottleneck starting from original image patch with a series of convolutional layers with dropout and pooling. In the decoder section, we gain back original resolution by upsampling and deconvolutional layers. In between, there are skip connections to concatenate lower and higher order features and easier flow of gradients. The detailed architecture is shown in Table \ref{tab_disc_architecture}. Unless otherwise stated, we use dropout\cite{dropout} with keep probability of 0.8. Leaky Relu activation is used after every convolution with negative gradient of 0.2.
\subsection{Generator Architecture}
For realizing the generator, we follow the principles in \cite{dcgan}. First, the 100D $z$ vector is passed through a linear layer and reshaped to a spatial resolution of $W/8 \times W/8$, where, $W \times W$ is the input patch resolution to the discriminator(W = 48 in our case). Then, we follow up with three transposed convolutional layers(also commonly known as deconvolutional layers) \cite{fcn} to increase resolution by 2X in each step to finally reach $W \times W$. Each layer is followed by Relu non linearity except the last layer which used tanh non linearity to scale output values in the range [-1, 1].
\subsection{Optimization}
We use mini batch stochastic gradient descent optimization with Adam optimizer\cite{adam} to train both generator and discriminator network. Learning rate for both the network are set to 10$^{-4}$. Batch size is kept at 64 and training usually progresses for 50 epochs in about 10 hours.
\begin{table}[!t]
\centering
\caption{Comparison of AUC on DRIVE dataset between our previous \cite{cvpr} and current method with additional unsupervised loss, $L_{unsup}$ (Eq. \ref{eq_unsupervised}). Here we use center pixel(CP) model of \cite{cvpr}.}
\begin{tabular}{l|llll}\hline
Method & \multicolumn{4}{c}{Annotated Patches} \\\hline
       & 0.5K     & 1K    & 2K    & 3K    \\
Lahiri \textit{et al.}\cite{cvpr}   & 0.82    & 0.84   & 0.85    & 0.81     \\
Proposed(CP)   & 0.86    & 0.88    & 0.89     & 0.90     \\\hline
\end{tabular}
\label{tab_poc_cvpr}
\end{table}
\section{Experiments}
\subsection{Datasets and Preprocessing}
We conduct experiments on DRIVE \cite{drive} and STARE\footnote{Available at: \href{http://cecas.clemson.edu/~ahoover/stare/}{http://cecas.clemson.edu/~ahoover/stare/}} datasets. DRIVE dataset has a clear demarcation of training and test set with 20 images in each category. Such breakup is not provided on STARE. Following recent practice \cite{tmi}, we follow a 1-held-out strategy, where we randomly select 1 image for testing and remaining 19 as train set. Results reported on STARE are average of 20 such trials.
\par The retinal images were converted to gray scale. It has been shown in \cite{lahiri2016deep} that the green channel in color fundus imaging is most discriminative in segmenting blood vessels. Following this, the green channel is given more weight in RGB to gray scale conversion. The contrast of the fundus images are improved using Contrast Limited Adaptive Histogram Equalization (CLAHE) and effect of non-uniform illumination is mitigated. Further, Gamma adjustment improves segmentation performance. Patches of resolution, 48$\times$48 are then extracted from the images.
\subsection{Benefit of Unsupervised Loss}
One of the major extensions in this paper over our prior work \cite{cvpr} is the addition of unsupervised loss, $L_{unsup}$, Eq. \ref{eq_unsupervised}. \textit{`But, does unsupervised loss help?'}  To investigate this, we first experimented with center pixel prediction method and exact same network architecture as in \cite{cvpr} on DRIVE. From Table \ref{tab_poc_cvpr}, it can be seen that addition of $L_{unsup}$ significantly ($p$ value $< 10^{-5}$) outperforms our previous framework consistently at different levels of annotation budget.
\subsection{GAN Hacks}
\label{sec_gan_hacks}
Finding Nash Equlibrium in a zero-sum minmax game such as in GANs is difficult (often resulting in oscillations) with stochastic gradient descent updates. This is a burning issue within GAN community. In this section, we present a detailed ablation study on various aspects of stabilizing GAN training with a basic U-Net as a baseline. Since The U-Net model is an essential component of many recent medical imaging applications, our findings in this section can serve as a guideline for any GAN based application which deploys U-Net at its core. In Table \ref{tab_gan_hacks} we report the AUC on DRIVE test set by training with 1K labeled samples with Feature Matching and vanilla GAN and also comparing the efficacy of different GAN stabilization techniques. Similar trends were also observed for STARE.
\begin{table}[!t]
\centering
\caption{Ablation study of two genres of GAN training (Feature Matching v/s Vanilla GAN) for semi supervised learning. Results show AUC on DRIVE with 1K labeled patches with U-Net architecture (Refer Table \ref{tab_disc_architecture} for architecture details). Different choices of pooling and kernel weight normalizations are reported. For feature matching, different options of matching feature statistics from convolutional layers, {$C_1$     , $C_3$     , $C_5$     , $C_7$ and $C_9$ } are explored while for vanilla GAN, domain prediction (real/fake) is taken from the last Softmax layer, $C_{10}$. }
\begin{tabular}{cllll|c}\hline
\multicolumn{5}{c|}{Feature Matching\cite{improved}} & Vanilla GAN\cite{goodfellow2014generative} \\\hline\hline
$C_1$     & $C_3$     & $C_5$     & $C_7$     & $C_9$    & $C_{10}$      \\\hline
\multicolumn{5}{c|}{Max Pool}       &                  \\
0.66   & 0.68  & 0.72   & 0.70   & 0.69  & 0.62    \\\hline\hline
\multicolumn{5}{c|}{Average Pool}   &                  \\
0.81   & 0.83   & 0.80  & 0.84  & 0.77 & 0.75    \\\hline\hline
\multicolumn{5}{c|}{Instance Norm + Average Pool}            &        \\
0.84   & 0.86   & 0.87   & 0.87  & 0.82 & 0.79     \\\hline\hline
\multicolumn{5}{c|}{Weight Norm + Average Pool}        &              \\
0.87  & 0.89  & 0.86   & 0.92   & 0.89 & 0.84    \\\hline
\end{tabular}
\label{tab_gan_hacks}
\end{table}

\subsubsection{Max Pool v/s Average Pool}
In an encoder-decoder architecture like the U-Net, it is common to use Max Pool operations for spatial reduction of intermediate feature layers. This results in sparse gradient operations which have been shown to hamper GAN training\cite{dcgan}. Specifically, a Max Pool operator, $M^{W \times W}(\cdot)$, operating on a receptive field of $W \times W$ resolution of a given feature map location, $F(x,y)$, results in finding the max value in $W \times W$ neighborhood.

$M^{W \times W}(F(x,y)) = max \{F(x - i, y-j) | $
\begin{equation}
i, j \in \{-W/2, -W/2 + 1,.., W/2\} \}
\label{eq_maxpool}
\end{equation}
Instead of using Max Pool, we benefited by using Average Pooling, which also achieves spatial reduction but with dense gradient operations. In line in notations with Eq. \ref{eq_maxpool}, we define Average Pool operator, $A^{W \times W}(\cdot)$ as,
\begin{equation}
A^{W \times W}(F(x,y)) = \frac{1}{W^2} \sum_{i=-W/2}^{W/2}\sum_{j=-W/2}^{W/2} F(x - i, y-j)
\end{equation}
In Table \ref{tab_gan_hacks} we see that Average Pool results in drastic improvement of performance compared to Max Pool. This observation holds true for training of both Feature Matching and vanilla variants of GAN. Based on our findings, we thus recommend the use of Average Pool over Max Pool in U-Net like architectures while training GANs.\\
\subsubsection{Normalization}
Normalization of intermediate activations/weights play a decisive role in success of training GANs. With the onset of `DCGAN' \cite{dcgan}, BatchNormalization (BN) \cite{batchnorm} has become the de facto choice of normalizing weights of a deep network for GAN training. While BN indeed speeds of training of GANs, recent works, specially in the domain of style transfer, recommends the use of Instance Normalization (IN) \cite{instancenorm} for better training of GANs. Our initial experiments also manifested better efficacy of IN over BN and thus in Table \ref{tab_gan_hacks} we report performance of models trained with IN + Average Pooling. IN + Max Pool did not show any significant improvement over only Max Pool which bolsters the fact that sparse gradient operations such as Max Pool are detrimental for GAN training. We further improved the performance by adopting the recent Weight Normalization (WN) technique proposed by Salimans \textit{et al}. \cite{weightnorm}\footnote{Implementation available at: \href{https://github.com/TimSalimans/weight\_norm}{https://github.com/TimSalimans/weight\_norm}}. For a linear layer,
\begin{equation}
\mathbf{y = W^TX + b},
\end{equation}
where $\mathbf{x} \in \mathbb{R}^n$, $\mathbf{y} \in \mathbb{R}^m$, $\mathbf{W} \in \mathbb{R}^{n\times m}$, WN re-parametrizes $\mathbf{W}$ with $\mathbf{V} \in \mathbb{R}^{n\times m}$ and a trainable scalar, $g \in \mathbb{R}^m$ according to,
\begin{equation}
\mathbf{w_i} = \frac{\mathbf{g_i}}{||\mathbf{v_i}||_2} \cdot \mathbf{v_i}.
\end{equation}
As shown in \cite{weightnorm}, decoupling of norm of the weight vector, $g$, from the direction of the weight vector, $\frac{\mathbf{v}}{||v||}$, helps in faster(and better) convergence of stochastic gradient descent optimization. Our GAN training also benefited using WN. In Table \ref{tab_gan_hacks} we report performance with WN + Average Pooling; this combination gives the best performance across all our experiments and unless otherwise stated, this should be taken as the default setting wherever we are using structured prediction with U-Net for all our further discussion on experiments. \\
\subsubsection{Selecting layer(s) for Feature Matching}
\begin{figure*}[!t]
\centering
\includegraphics[scale = 0.4]{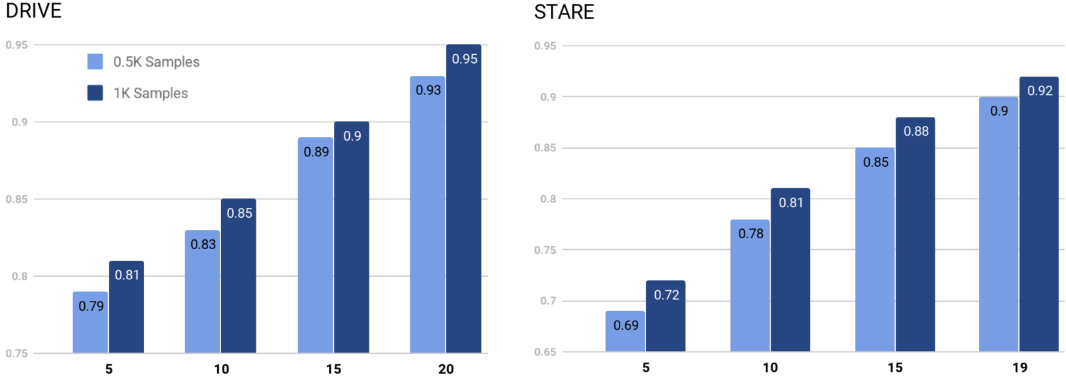}
\caption{AUC (Y-axis) on DRIVE and STARE datasets at different budgets (0.5K, 1K) of annotated samples conglomerated from different number of training images (X-axis). It is observed that models trained with a lower budget of annotated patches but sampled from a larger pool of images results in better performance than models trained with more annotations on smaller pool of images. This recommends diversity of samples over quantity of samples specifically while designing low annotation deep learning models.}
\label{fig_num_images}
\end{figure*}

\begin{figure*}[!h]
\centering
\begin{minipage}[t]{0.96\textwidth}
\includegraphics[width=\linewidth]{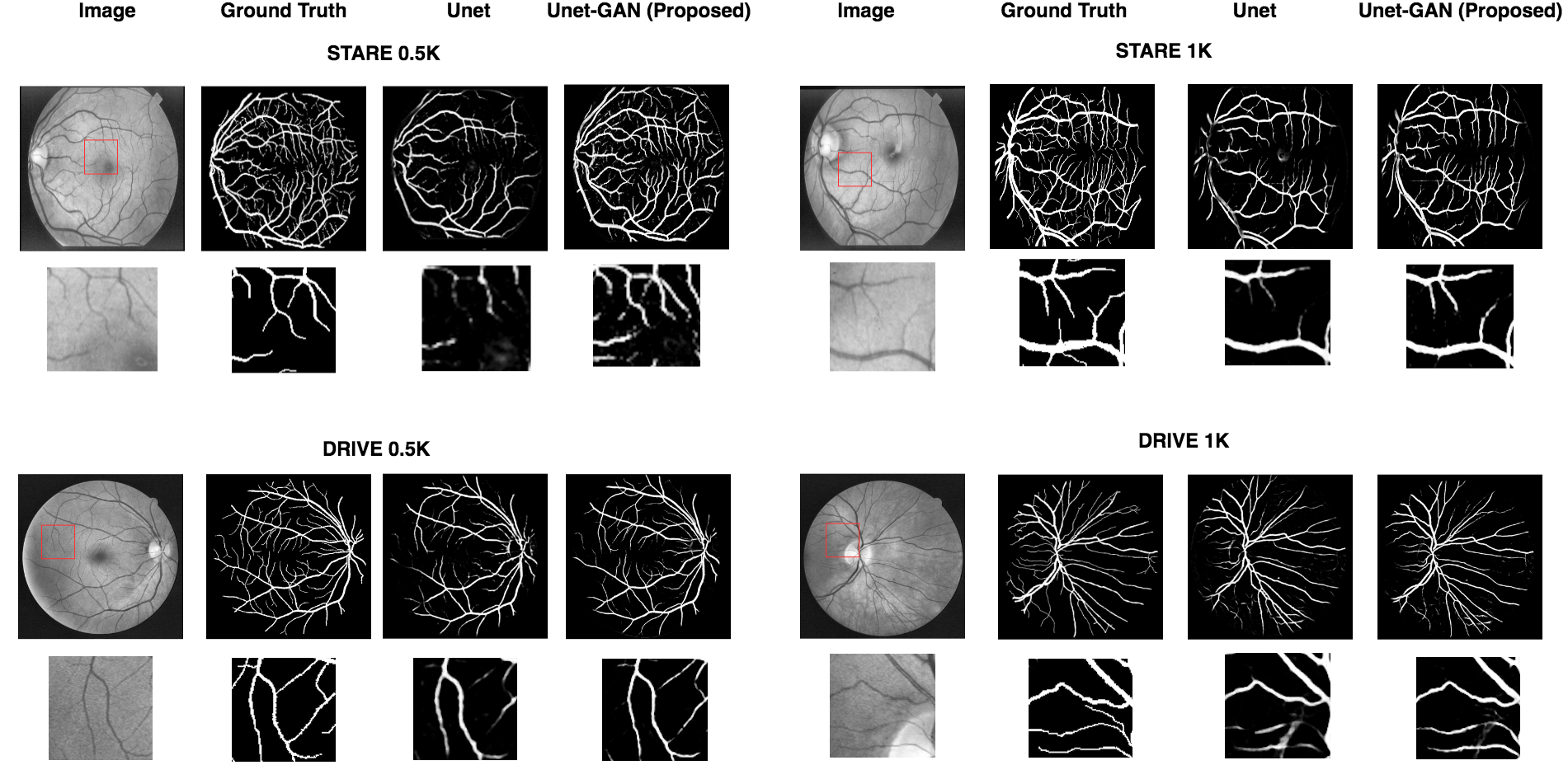}
\end{minipage}
\caption{Some sample visualizations of segmented vessels on DRIVE and STARE dataset at 0.5K and 1K patch annotation budget. For each figure, we also show a zoomed up section. Even with 0.5K samples, our method appreciable efficacy at segmenting finer vessels compared to supervised U-Net model. The effect is more pronounced on STARE dataset which consists of data from patient group with various opthalmic disorders.}
\label{fig_visual_comparison}
\end{figure*}

In their original implementation, \footnote{Available at: \href{https://github.com/openai/improved-gan}{https://github.com/openai/improved-gan}} Salimans \textit{et al.} \cite{improved} used the penultimate layer of the discriminator for matching features from a batch of real and fake samples. We hypothesize that for low level vision tasks, matching features from such deeper layers of a network is not a prudent approach. For cases in which the end task is simple classification, such as in \cite{improved}, it makes sense to only focus on higher order features from deeper parts of the network. Features essential for classification are agnostic to local perturbations. But in our case, the fully convolutional discriminator is responsible for semantic segmentation - assigning class label to each pixel of a patch. This requires low level information along with high level features. In fact, our initial experiments with feature matching on the penultimate layer of discriminator yielded the worst AUC performance. For initial investigation, we trained a separate model with a different layer (Refer to Table \ref{tab_disc_architecture} for details of network layer) selected from the discriminator to adapt. In Table \ref{tab_gan_hacks} we report the AUC values on DRIVE dataset for those models. It appears that selecting the extremely shallow or deep layer hurts the performance. It is prudent to match intermediate layers for our low level vision task. As a step further, we experimented with matching the layers at the points of concatenation, Con$_1$ and Con$_2$. We achieved best performance when we match features at Con$_1$ layer which is a combination of layers C$_4$ and U$_1$ (upsampled from C$_5$). Henceforth, our results(Tables \ref{tab_compare_all_DRIVE}, \ref{tab_compare_all_STARE}, Figures \ref{fig_num_images}, \ref{fig_visual_comparison}) will be reported with this setting. It is also be seen that training the generator with the original GAN formulation \cite{goodfellow2014generative} hurts semi-supervised performance and thus moving forward we experimented with only feature matching paradigm. The proof of concepts learnt so far on DRIVE were also extended on STARE dataset experiments, unless otherwise stated.
\subsection{More Patches or More Images ?}
\label{sec_diversity}
Till now, we have been referring only to the total number of patches as the annotation budget constraint. However, when working with such low annotation setting, it is important to investigate the importance of number of annotations and diversity of annotations. For example, we can have a budget of 500 annotated samples; these 500 patches can be taken from a single image in worst case or equally sampled from all training set images in best case. In Fig. \ref{fig_num_images} we compare the performance of our model trained with two budgets - 0.5K and 1K. Annotations are sampled uniformly from different subsets of entire training set. It is observed that at a given budget of annotation, if we sample from more number of training images, then performance improves. This is especially evident in experiments performed on the STARE dataset which has more number of patient samples compared to DRIVE. Another interesting observation is that models trained with lower annotation budget but sampled from a larger pool of training data tends to perform better than models trained on higher annotation budget but from smaller pool of images. For example, on STARE, we achieve an AUC of 0.85 with 0.5K annotations sampled from 15 images compared to 0.81 achieved by training on 1K annotation from 10 images. Similar trends can be found on DRIVE dataset. These observations suggest that, for a given annotation budget, one should try to collect more images and perform less annotation per image.
\subsection{Comparison with State-of-the-art}
\label{sec_state_of_art}
In Tables \ref{tab_compare_all_DRIVE} and \ref{tab_compare_all_STARE} we compare performance of our U-Net GAN model with the vanilla U-Net model \footnote{Implementation adapted from \href{https://github.com/orobix/retina-unet}{https://github.com/orobix/retina-unet}}. At full supervision with 60K labeled samples, the vanilla U-Net achives AUC of 0.97 on DRIVE and 0.96 on STARE and thus U-Net serves as a very strong baseline for supervised training. At very low number of annotated patches, our model consistently outperforms U-Net across both datasets. Also, we gain distinct gain over our previous semi-supervised framework \cite{cvpr}. We also compared against two contemporary benchmark supervised benchmark models of \cite{avijit} and \cite{tmi} and achieved consistent gains at different levels of annotation on both datasets. The current work thus sets up a new benchmark for such low annotation retinal vessel segmentation across two real life fundus datasets.
\begin{table}[!t]
\centering
\caption{Comparison of competing supervised and semi supervised methods on DRIVE dataset.}
\begin{tabular}{llllll} \hline

Genre           & Method & \multicolumn{4}{c}{Annotated Patches} \\ \hline\hline
                &        & 0.5K   & 1K  & 3K  & 10K  \\\hline
      & Dasgupta \textit{et al.}\cite{avijit}  & 0.85  & 0.87  & 0.89  & 0.92   \\
       Supervised         & Liskowski \textit{et al.}\cite{tmi}    & 0.83  & 0.84  & 0.87  & 0.92   \\
                & U-Net   & 0.89  & 0.90  & 0.92  & 0.95   \\\hline
Semi Supervised & Lahiri \textit{et al.} \cite{cvpr} & 0.82  & 0.84  & 0.85  & 0.93   \\
                & Proposed (SP)   & 0.92  & 0.94  & 0.96  & 0.97  \\\hline
\end{tabular}
\label{tab_compare_all_DRIVE}
\end{table}
\begin{table}[!h]
\centering
\caption{Comparison of competing supervised and semi supervised methods on STARE dataset.}
\begin{tabular}{llllll} \hline

Genre           & Method & \multicolumn{4}{c}{Annotated Patches} \\ \hline\hline
                &        & 0.5K   & 1K  & 3K  & 10K  \\\hline
      & Dasgupta \textit{et al.}\cite{avijit}  & 0.82  & 0.84  & 0.87  & 0.91   \\
       Supervised         & Liskowski \textit{et al.}\cite{tmi}    & 0.84  & 0.86  & 0.89  & 0.93   \\
                & U-Net   & 0.86  & 0.89  & 0.90  & 0.94   \\\hline
Semi Supervised & Lahiri \textit{et al.} \cite{cvpr} & 0.80  & 0.81  & 0.83  & 0.90   \\
                & Proposed (SP)   & 0.90  & 0.92  & 0.94  & 0.96  \\\hline
\end{tabular}
\label{tab_compare_all_STARE}
\end{table}

\section{Conclusion and Discussion}
In this paper we extended our previous work \cite{cvpr} on semi supervised segmentation of retinal vessels from fundus images. We showed how unlabeled training samples can be leveraged via unsupervised adversarial loss function leading to final boost of segmentation performance. Our rigorous experiments recommend a series of best practices while training an encoder-decoder like architecture in a GAN framework. It was consistently seen across both datasets, then in the regime of low annotation space, GAN based semi-supervised learning performs better than state-of-the-art supervised models. Our work thus opens up new opportunites to leverage deep learning framework on domains where availability of annotated data is scarce. We made an important observation that diversity of annotation is more important than actual number of annotations. This observation is particularly promising since getting medical imaging data is still easier with help of para medics than annotating them with experts. Since, we made no assumption on the distribution of underlying data, our findings should be seamlessly applicable for other medical imaging domains as well.
\section{Acknowledgement}
The authors would like to thank the repository owners of \cite{unet_git, dcgan_git} for open sourcing their code on which we built upon. Avisek is funded by a Google PhD Fellowship in Machine Perception.

\bibliographystyle{ieee}
\bibliography{egbib}
\end{document}